\documentclass[11pt,twoside]{article}
\usepackage{eacl2003}
\usepackage{times}
\usepackage{latexsym}

\title{A Flexible Pragmatics-driven Language Generator for Animated Agents}

\author{Paul Piwek \\ {\sc itri} -- Information Technology Research Institute \\ University of Brighton \\ {\tt Paul.Piwek@itri.bton.ac.uk}}

\date{}

\begin{document}

\maketitle

\begin{abstract}

This paper describes the {\sc neca mnlg}; a fully implemented Multimodal Natural Language Generation module. The {\sc mnlg} is deployed as part of  the {\sc neca} system which generates dialogues between animated agents. The generation module supports the seamless integration of full grammar rules, templates and canned text. The generator takes input which allows for the specification of syntactic,  semantic and pragmatic constraints on the output.

\end{abstract}

\section{Introduction}

This paper introduces the {\sc neca} {\sc mnlg}; a Multimodal Natural Language Generator. It has been developed in the context of the {\sc neca} system.\footnote{{\sc neca} stands for `Net Environment for Embodied Emotional Conversational Agents' and is an {\sc eu-ist} project.} The {\sc neca} system generates dialogue scripts for animated characters. A first demonstrator in the car sales domain ({\sc eShowroom}) has been implemented. It allows a user to browse a database of cars, select a car, select two characters and their attributes, and subsequently view an automatically generated film of a dialogue about the selected car. The demonstrator takes the following input:

{\small
\begin{itemize}
\setlength{\itemsep}{-0.1cm}
\item A database with facts about the selected car (maximum speed, horse power, etc.). 
\noindent
\item A database which correlates facts with value judgements.
\noindent
\item Information about the characters: 1. Personality traits such as extroversion and agreeableness. 2. Personal preferences concerning cars (e.g., a preference for safe cars). 3. Role of the character (seller or customer).
\end{itemize}
}
\noindent
This input is processed in a pipeline that
consists of the following modules in this order: 

{\small
\begin{itemize}
\setlength{\itemsep}{-0.1cm}
\item A {\sc Dialogue Planner}, which produces an abstract 
description of the dialogue (the dialogue plan). 
\item A {\sc multi-modal Natural Language Generator} which specifies 
linguistic and non-linguistic realizations for the dialogue acts 
in the dialogue plan. 
\item A {\sc Speech Synthesis Module}, which adds information for Speech. 
\item A {\sc Gesture Assignment Module}, which controls the temporal 
coordination of gestures and speech.  
\item A {\sc player}, which plays the animated characters and the corresponding speech sound files. 
\end{itemize}
}
\noindent
Each step in the pipeline adds more concrete information to the
dialogue plan/script until finally a player can render it. A single
{\sc xml} compliant representation language, called {\sc rrl}, has been developed for representing the Dialogue Script at its various stages of completion (Piwek et al., 2002).

In this paper, we describe the requirements for the {\sc neca mnlg}, how these have been translated into design solutions and finally some of aspects of the implementation. 

\section{Requirements}

The requirements in this section derive primarly from the use case of the {\sc neca} system. We do, however, try to indicate in what respects these requirements transcend this specific application and are desirable for generation systems in general.
\\
\\
\noindent
{\small
{\sc Requirement} {\bf 1}: {\it The linguistic resources of the generator should support seamless integration of canned text, templates and full grammar rules}. 
}
\\
\\
\noindent
In the {\sc neca} system, the dialogue planner creates a dialogue plan consisting of (1) a description of the participants, (2) a characterization of the common ground at the outset of the dialogue in terms of Discourse Representation Theory (Kamp and Reyle, 1993) and (3) a set of dialogue acts and their temporal ordering. For each dialogue act, the type, speaker, set of addressees, semantic content, what it is a reaction to (i.e., its rhetorical relation to other dialogue acts), and emotions of the speaker can be specified. The amount of information which the dialogue planner actually provides for each of these attributes varies, however, per dialogue act: for some dialogue acts, a full semantic content can be provided --in the form of a Discourse Representation Structure-- whereas for other acts, no semantic content is available at all. Typically, the dialogue planner can provide detailed semantics for utterances whose content is covered by the domain model (e.g., the car domain) whereas this is not possible for utterances which play an important role in the conversation but are not part of the domain model (e.g., greetings). This state of affairs is shared with most real-world applications. 

Since generation by grammar rules is primarily driven by the input semantics, for certain dialogue acts full grammar rules cannot be used. These dialogue acts may be primarily characterized in terms of their, possibly domain specific, dialogue act type (greeting, refusal, etc.). Thus, we need a generator which can cope with both types of input, and map them to the appropriate output. Input with little or no semantic content can typically be dealt with through templates or canned text, whereas input with fully specified semantic content can be dealt with through proper grammar rules. Summarizing, we need a generator which can cope with (linguistic) resources that contain an arbritary combination of grammar rules, templates and canned text.
\\
\\
\noindent
{\small
{\sc Requirement} {\bf 2}: {\it The generator should allow for combinations of different types of constraints on its the output, such as syntactic, semantic and pragmatic constraints} 
}
\\
\\
\noindent
In the {\sc neca} project the aim is to generate behaviour for animated agents which simulates affective situated face-to-face conversational interaction. This means that the utterances of the agents have to be adapted not only to the content of the information which is exchanged but also to many other properties of the interlocutors, such as their emotional state, gender, cultural background, etc. The generator should therefore allow for such parameters to be part of its input. 
\\
\\
\noindent
{\small
{\sc Requirement} {\bf 3}: {\it The generator should be sufficiently fast to be of use in real-world applications}
}
\\
\\
\noindent
The application in which our generator is being used is currently fielded as part of a net-environment. The application will be evaluated with users through online questionnaires which are integrated in the application and analysis of log files (to answer questions such as `Do users try different settings of the application?', etc. See Krenn et al., 2002). Therefore, the generator will have to be fast in order for it not to negatively affect the user experience of the system.

\section{Design Solutions}

The {\sc neca} {\sc mnlg} adopts the conventional pipeline architecture for generators (Reiter and Dale, 2000). Its input is a {\sc rrl}  dialogue plan. This is parsed and internally represented as a {\sc Profit} typed feature structure (Erbach, 1995). Subsequently, the dialogue acts in the plan are realized in accordance with their temporal order. For each act, first a deep syntactic structure is generated. The deep structure of referring expressions is dealt with in a separate module, which takes the common ground of the interlocutors into account. Subsequently, lexical realization (agreement, inflection) and punctuation is performed. Finally, turn-taking gestures are added and the output is mapped back into the {\sc rrl} {\sc xml} format. 

Here let us concentrate on our approach to the generation of deep syntactic structure and how it satisfies the first two requirements. The input to the {\sc mnlg} is a node (i.e., feature structure) stipulating the syntactic type of the output (e.g., sentence: \verb"<s"), semantics and further information on the current dialogue act in {\sc Profit}:\footnote{That is, {\sc Prolog} with some sugaring for the representation of feature structures. Feature structures are also used in the {\sc fuf/surge} generator. It is different from the {\sc neca mnlg} in that it takes as input thematic trees with content words. Furthermore, it allows for control annotations in the grammar and uses a special interpreter for unification, rather than directly {\sc Prolog}. See {\tt http://www.cs.bgu.ac.il/surge/}.}

{\footnotesize
\begin{verbatim}
(<s &
 sem!drs([c_27],
         [type(c_27,prestigious),
          arg1(c_27,x_1)])&
 currentAct!speaker!
    (name!john &
     polite!yes & ...)
)
\end{verbatim}
}

\noindent
Thus various types of information are combined within one input node. Generation consists of taking the input node and using it to create a tree representation of the output. For this purpose, the {\sc mnlg} tries to match the input node with the mother node of one of the trees in its tree repository. This tree repository   contains trees which can represent proper grammar rules, templates and canned text. Matching trees might in turn have incomplete daughter nodes. These are recursively expanded by matching them with the trees in the repository, until all daughters are complete. 

A daughter node is complete if it is lexically realized (i.e., the attribute \verb"form" of the node has a value) or it is of the type \verb"<np" and the semantics is an open variable. In the latter instance, the node is expanded in a separate step by the referring expressions generation module. This module finds the discourse referent in the common ground which binds the open variable and constructs a description of the object in question. The description is composed of the properties which the object has according to the common ground, but can also be empty if the object is highly salient. The module is based on the work of Krahmer and Theune (2002). The (empty) description is mapped to a deep syntactic structure using the tree repository. Lexicalization subsequently yields expressions such as `it' (empty descriptive content) or, for instance, `the red car'. 

Let us return to the tree repository and illustrate how templates  and rules can be represented uniformly. The representation of a tree is of the form \verb"(Node,[Tree1,Tree2,...])", where the list of trees can be empty, yielding a tree consisting of one node: \verb"(Node,[])". The following is a template for dialogue acts of type \verb"greeting" with no semantic content and a polite speaker.  

{\footnotesize
\begin{verbatim}
(<s &
 currentAct!
   (type!greeting &  
    speaker!polite!"yes" &
    speaker!name!Speaker) & 
 sem!"none",     
       [(<s & form!"hello!",[]),
        (<fragment & 
         form!"My name is",[]),
        (<np & 
         sem!concept(Speaker),[])
        ]).
\end{verbatim}
}

\noindent
This is a template for the text `Hello! My name is {\sc Speaker}'.
Where {\sc Speaker} is a variable which is bound to the name of the speaker of the utterance. The noun phrase (\verb"<np") for this name is generated by the referring expression generation module. The following is a tree representing a grammar rule of the familiar type {\it S} $\rightarrow$ {\it NP VP}:

{\footnotesize
\begin{verbatim}
(<s &
 currentAct!type!statement &
 currentAct!CA &
 argGap!ArgGap &
 auxGap!AuxGap &        
 sem!drs(_,[negation(
              drs(_,
                  [type(E,Type)
                   arg1(E,X)|R]))]
                  ),       
        [(<np &
          currentAct!CA &
          sem!X,[]),
         (<vp &
          argGap!ArgGap &
          auxGap!AuxGap &
          negated!<true & 
          sem!drs(_,[type(E,Type)|R]) &
          currentAct!CA,_)
         ]).
\end{verbatim}
}

\noindent
Note that this rule applies to an input node whose semantic content contains a negation. The negation is passed on to the {\it VP} subtree via the feature \verb"negated". The attributes \verb"argGap" and \verb"auxGap" allow us to capture unbounded dependencies via feature perlocation. Our use of trees is related to the Tree Adjoining Grammar approach to generation (e.g., Stone and Doran, 1997).\footnote{Their generation algorithm is, however, very different from the one proposed here. Whereas they propose an integrated planning approach, we advocate a very modular system, supporting fast generation. Moreover, by using features for unbounded dependencies we do not require the adjunction operation, which is incompatible with our topdown generation approach. We follow Nicolov et al. (1996), who also use {\sc tag}, in their commitment to flat semantics. Their generator does, however, not take pragmatic constraints into account.}  

The value of the attribute \verb"currentAct" is passed on from the mother node to the daughter nodes. Thus any pragmatic information (personality, politeness, emotion, etc.) is passed on through the tree and can be accessed at a later stage, for instance, when lexical items are selected.

\section{Implementation}

The {\sc neca mnlg} has been implemented in {\sc prolog}. The output is in the form of an {\sc rrl} {\sc xml} document. Table 1 provides a sample of the response times of the compiled code running on a Pentium {\sc iii} Mobile 1200 Mhz with Sicstus 3.8.5 {\sc prolog}. We timed the complete generation process from parsing the {\sc xml} input to producing {\sc xml} output, including generation of deep syntactic structure, referring expressions, turn taking gestures (not discussed in this paper), etc.

\begin{table}[h]
\centering
{\footnotesize
\begin{tabular}{| c | c | c | c | }
\hline
input & \# acts & $=1$ & $\leq10$ \\ 
\hline  
{\sc a} & 19 & 0.230s & 0.741s \\
{\sc b} & 22 & 0.290s & 0.872s \\
{\sc c} & 23 & 0.290s & 0.801s \\
{\sc d} & 31 & 0.431s & 1.372s \\
\hline
\end{tabular}
}
\label{times}
\caption{Response Times of the {\sc mnlg}}
\end{table}

\noindent
The results show generation times for entire dialogues and according to whether the generator was asked to produce exactly one solution or select at random a solution from a set of at most ten generated solutions (the latter strategy was implemented to obtain more variation in the generator output). On average for $=1$ the generation time for an individual dialogue act is almost $\frac{1}{100}$ of a second. For $\leq10$ it is $\frac{4}{100}$ of a second. The generator uses a repository of 138 trees (including the two examples given above). The repository has been developed for and integrated into the {\sc eShowroom} system which is currently being fielded. A start is being made with porting the {\sc mnlg} to a new domain and documentation is being created to allow our project partners to carry out this task. We hope that our efforts will contribute to addressing a challenge expressed in (Reiter, 1999): ``We hope that future systems such as {\sc stop} will be able to make more use of deep techniques, because of advances in linguistics and the development of reusable wide-coverage {\sc nlg} components that are robust, well-documented and well engineered as software artifacts.''

In our view the best way to approach this goal is by providing a framework which allows for the flexible integration of shallow and deep generation, thus making it possible that in the course of various projects, deep analyses can be developed alongside the shallow solutions which are difficult to avoid altogether in software development projects, due to the pressure to deliver a {\it complete} system within a certain span of time.  

\subsection*{Acknowledgements} This research is supported by the {\sc eu} Project {\sc neca ist-}2000-28580. For comments and discussion thanks are due the {\sc eacl} reviewers and my colleagues in the {\sc neca} project.  

\subsection*{References}

{\small
\begin{description}
\setlength{\itemsep}{-0.1cm}

\item[] Gregor Erbach. 1995. {\sc Profit} {\it 1.54 user's guide}. University of the Saarland, December 3, 1995.

\item[] Hans Kamp and Uwe Reyle. 1993. {\it From Discourse
to Logic}. Kluwer, Dordrecht.

\item[] Emiel Krahmer and Mari\"et Theune. 2002. Efficient context-sensitive generation of referring expressions. In: Kees Van Deemter and Rodger Kibble (eds.), {\it Information Sharing}, {\sc csli}, Stanford.

\item[] Brigitte Krenn, Erich Gstrein, Barbara Neumayr and Martine Grice. 2002. What can we learn from users of avatars in net environments?. In: {\it Proc. of the AAMAS workshop ``Embodied conversational agents - let's specify and evaluate them!''}, Bologna, Italy.

\item[] Nicholas Nicolov, Chris Mellish \& Graeme Ritchie. 1996. Approximate Generation from Non-Hierarchical Representattions, {\it Proc. 8th International Workshop on Natural Language Generation}, Herstmonceux Castle, UK.

\item[] Paul Piwek, Brigitte Krenn, Marc Schr\"oder, Martine Grice, Stefan Baumann and Hannes Pirker. 2002. RRL: A Rich Representation Language for the Description of Agent Behaviour in NECA. {\it Proc. of the AAMAS workshop ``Embodied conversational agents - let's specify and evaluate them!''}, Bologna, Italy.

\item[] Ehud Reiter. 1999. Shallow vs. Deep Techniques for handling Linguistic Constraints and Optimisations. {\it Proc. of KI-99 Workshop "May I speak freely"}.

\item[] Ehud Reiter and Robert Dale. 2000. {\it Building natural language generation systems}. Cambridge University Press, Cambridge.

\item[]Matthew Stone and Christy Doran. 1997. Sentence Planning as Description Using Tree-Adjoining Grammar. {\it Proc. ACL 1997}, Madrid, Spain.

\end{description}
}

\end{document}